\documentclass[a4paper]{article}

\usepackage{setspace}
\usepackage{authblk}
\usepackage{enumitem}

\usepackage{makeidx}
\usepackage{a4wide}
\usepackage[pdftex]{graphicx}
\usepackage{dsfont}
\usepackage{amsmath}
\usepackage{amssymb} 
\usepackage{listings}
\usepackage{algorithm}
\usepackage{gensymb}

\usepackage{cite}
\usepackage{url}
\usepackage{epsfig}
\usepackage{color}
\usepackage{psfrag}

\usepackage{pbox}

\usepackage[english]{babel}
\usepackage[usenames,dvipsnames]{xcolor}

\usepackage{rotating}
\usepackage{rotate}
\usepackage{lscape}

\usepackage{multirow}

\usepackage[colorinlistoftodos]{todonotes}
\usepackage{xcolor}

\usepackage{dcolumn}
\usepackage{bm}
\usepackage{float}
\floatstyle{plain}
\restylefloat{figure}

\usepackage[T1]{fontenc}
\usepackage[utf8]{inputenc}
\usepackage{tabularx,ragged2e,booktabs,caption}
\newcolumntype{C}[1]{>{\Centering}m{#1}}

\def\lim{\mathop{\rm lim}}

\makeatletter
\newcommand*{\rom}[1]{\expandafter\@slowromancap\romannumeral #1@}

\usepackage{tikz}
\usetikzlibrary{shapes,arrows,chains}

\usepackage{soul}  

\setulcolor{red} 
\setstcolor{red} 
\sethlcolor{yellow} 

\usepackage{amsthm}

\title{ A clarification of misconceptions, myths and desired status of artificial intelligence}

\author[1,2]{\small Frank Emmert-Streib}
\author[2]{\small Olli Yli-Harja}
\author[3]{\small Matthias Dehmer}
\affil[1]{\tiny Predictive Society and Data Analytics Lab, Faculty of Information Technology and Communication Sciences, Tampere University, Tampere, Finland \thanks{frank.emmert-streib@tuni.fi}}
\affil[2]{Institute of Biosciences and Medical Technology, Tampere University of Technology, Tampere, Finland}
\affil[3]{Department of Mechatronics and Biomedical Computer Science, UMIT, Hall in Tyrol, Austria}

\begin{document}

\pgfdeclarelayer{marx}
\pgfsetlayers{main,marx}
\providecommand{\cmark}[2][]{%
  \begin{pgfonlayer}{marx}
    \node [nmark] at (c#2#1) {#2};
  \end{pgfonlayer}{marx}
  }
\providecommand{\cmark}[2][]{\relax}

{\setstretch{1.0}

\maketitle

}

\begin{abstract}

The field artificial intelligence (AI) has been founded over 65 years ago. Starting with great hopes and ambitious goals the field progressed though various stages of popularity and received recently a revival in the form of deep neural networks. Some problems of AI are that so far neither 'intelligence' nor the goals of AI are formally defined causing confusion when comparing AI to other fields. In this paper, we present a perspective on the desired and current status of AI in relation to machine learning and statistics and clarify common misconceptions and myths. Our discussion is intended to uncurtain the veil of vagueness surrounding AI to see its true countenance.

\end{abstract}

\section{Introduction}

Artificial intelligence (AI) has a long tradition. The name AI was coined by McCarthy at the Dartmouth conference in 1956 starting a concerted endeavor that continues to date \cite{mccarthy2006proposal}. The initial focus of AI was on symbolic models and reasoning as search followed by the first wave of neural networks and expert systems \cite{rosenblatt1957perceptron,crevier1993ai,newel1976completer}. In the 1980s neural networks had a first return by invention of the back-propagation algorithm \cite{rumelharthinton1_1986} and in the 1990s research about intelligent agents received broad interest. Recently, big data became available and led to revival of neural networks in the form of deep neural networks \cite{lecun2015deep,hochreiter1997long}. 

AI has achieved great successes in many different fields including robotics, speech recognition, facial recognition, healthcare and finance   \cite{bahrammirzaee2010comparative,brooks_science1991,krizhevsky2012imagenet,hochreiter1997long,thrun2002robotic,yu2018artificial}. Given the breath of AI applications and the variety of methods used it is no surprise that seemingly simple questions, e.g., regarding the aims and goals of AI got obscured especially for those scientists who did not accompany the field since its initiation 65 years ago. For this reason, in this paper, we discuss the desired and current status of AI regarding its definition and provide a clarification for the discrepancy. Specifically, we provide a perspective on AI in relation to machine learning and statistics.

Our paper is organized as follows. In the next section, we discuss the desired and current status of artificial intelligence including the definition of 'intelligence' and strong AI. Then we clarify frequently encountered misconceptions about AI. Finally, we discuss characteristics of methods from artificial intelligence in relation to machine learning and statistics. The paper finishes with concluding remarks.

\section{What is artificial intelligence?}\label{sec.eai}

We begin our discussion by clarifying the meaning of artificial intelligence. We start by discussing definitions of 'intelligence' followed by informal characterizations of AI because the former problem will turn out to be currently unresolvable.

\subsection{Defining 'intelligence' in AI}

From the name 'artificial intelligence' it seems obvious that AI is dealing with an artificial - not natural - form of intelligence. Hence, defining 'intelligence' in a precise way will tell us what AI is about. Unfortunately, to this day there is no such definition. In \cite{legg2007universal} the difficulties encountered when attempting to provide such a definition are discussed and a formal measure is suggested. Interestingly, the authors start from several informal definitions of human intelligence to define machine intelligence formally. The resulting measure is given by
\begin{eqnarray}
\Upsilon(\pi) = \sum_{\mu \in E} 2^{-K(\mu)} V_{\mu}^{\pi}.
\end{eqnarray} 
Here $\pi$ is an agent, $K$ the Kolmogorov complexity function, $E$ the set of all environments, $\mu$ one particular environment, $2^{-K(\mu)}$ the algorithmic probability distribution over an environment and $V_{\mu}^{\pi}$ a value function. Overall, $\Upsilon(\pi)$ is called the {\it universal intelligence} of agent $\pi$ \cite{legg2007universal}. Informally, it gives a measure for the intelligence as the ability of an agent to achieve goals in a wide range of environments \cite{legg2007universal}.

A general problem with this definition is that its form is rather cumbersome and unintuitive, and its exact practical evaluation is not possible because the Kolmogorov complexity function K is not computable but requires approximation. A further problem is to perform intelligence tests because, e.g., a Turing test is insufficient.  

A good summary of the problem in defining 'intelligence' and AI is given in \cite{winston1984artificial} stating that "Defining intelligence usually takes a semester-long  struggle, and even after that I am not sure we ever get a definition really nailed down. But operationally speaking, we want  to make machines smart." In summary, there is currently no generally accepted definition of 'intelligence' neither tests that could be used to identify it reliably.

Despite this lack of a general definition of 'intelligence' there is a philosophical separation of AI systems based on this notion.  The so called {\it weak AI hypothesis} states that "machines could act as if they were intelligent" whereas the {\it strong AI hypothesis} asserts "that machines that do so are actually thinking (not just simulating thinking)" \cite{russell2016artificial}. Especially, the latter is very controversial and an argument against a strong AI is the Chinese room \cite{searle2008mind}. We would like to note that strong AI has been recently rebranded as {\it artificial general intelligence} (AGI) \cite{goertzel2007artificial}.

\subsection{Informal characterizations of AI}

Due to this lack of a general definition of 'intelligence' AI has been characterized informally from its beginnings. For instance, in \cite{winston1984artificial} it has been stated that "The primary goal of Artificial Intelligence is to  make machines  smarter. The secondary goals of Artificial Intelligence are to understand what intelligence is (the Nobel laureate purpose) and to make machines more useful (the entrepreneurial purpose)"; Kurzweil \cite{kurzweil1990age} noted that "The art of creating machines that perform functions that require intelligence when performed by people"; and Feigenbaum  \cite{feigenbaum1963artificial} said "artificial intelligence research is concerned with constructing machines (usually programs for general-purpose computers) which exhibit behavior such that, if it were observed in human activity, we would deign to label the behavior 'intelligent'." This reminds to a Turing test of intelligence.

Feigenbaum further specifies that "One group of researchers is concerned with simulating human information-processing activity, with the quest for precise psychological theories of human cognitive activity" and "A second group of researchers is concerned with evoking intelligent behavior from machines whether or not the information processes employed have anything to do with plausible human cognitive mechanisms" \cite{feigenbaum1963artificial}. Similar distinctions have been made in \cite{pomerol1997artificial,simon1969sciences}. Interestingly, the first point addresses a natural - not artificial - form of cognition showing that some scientists even cross the boundary of artificial to biological phenomena.

From this follows, that from its beginnings AI had high aspirations focusing on ultimate goals centered around intelligent and smart behavior rather than on simple questions as represented, e.g., by classification or regression problems. This means also that AI is not explicitly data-focused but assumes the availability of data that would allow the studying of such high-hanging questions. This is in contrast to data science which aims to extract the optimum on information contained in data set(s) possibly by applying more than one method \cite{em_jemmert2019defining}.

 \subsection{Current status}
 
From the above discussion is seems fair to assert that neither do we have a generally accepted, formal (mathematical) definition of 'intelligence' nor do we have one succinct informal definition of AI that would go beyond its obvious meaning. Instead, there are many different characterizations and opinions about what AI should be \cite{wang2006rigid}.

\section{Common misconceptions and myths}

In this section, we discuss some frequently encountered misconceptions about AI. In the following, we clarify some falsely made assumptions.

{\it AI aims to explain how the brain works}. No, because brains occur only in living (biological) beings and not artificial machines. Hence, the fields studying the molecular biological mechanisms of natural brains are neuroscience and neurobiology. If AI research may nevertheless contribute to this question in some way is unclear but so far no contribution has been made.

{\it AI methods work similar as brains}. No, although the most prominent methods of AI are called neural networks which are inspired by biological brains. Importantly, despite the name 'neural network' such models do not present physiological neural models because neither the neuron model nor the connectivity between the neurons is biologically plausible nor realistic. Specifically, a physiological model of a biological neuron is the Hodgkin-Huxley model \cite{hodgkinhuxley_1952} or the FitzHugh-Nagumo model \cite{nagumo_1962} and the connectivity is to date unknown. However, neither the connectivity structure of convolutional neural networks nor that of deep feedforward neural networks is biologically realistic.

{\it Methods from AI have a different purpose as methods from machine learning or statistics}. No, the general purpose of all methods from these fields is to analyze data. However, each field introduced different methods having different underlying philosophies. Specifically, the philosophy of AI is to aim at ultimate goals, which are possibly unrealistic, rather than to answer simple questions.

{\it AI is a technology}. No, AI is a methodology. That means the methods behind AI are (mathematical) learning algorithms that adjust the parameters of methods via learning rules. However, when implementing AI methods certain problems may require an optimization of the method in combination with computer hardware, e.g., by using a GPU, in order to improve the computation time it takes to execute a task. The latter combination may give the impression that AI is a technology but by downscaling a problem one can always reduce the hardware requirements demonstrating the principle workings of a method.

{\it AI makes computers think}. From a scientific point of view no, because similar to the problems defining 'intelligence' there is currently no definition of 'thinking'. Also thinking is in general associated with humans which are biological beings rather than artificial machines. In general, this point is related to the goals of strong AI and the counter argument by Searle \cite{searle2008mind}.

{\it Why appears AI more mythical than machine learning or statistics?} Considering the fact that both fields serve a similar purpose (see above) this is indeed strange. However, we think that the reason therefore is twofold. First, the vague definition of AI leaves much room for guesswork and wishful thinking and second, the high aspiration of AI enables speculations about ultimate or futuristic goals like 'making machines think' or 'making machines human-like'.

{\it Making machines to behave like humans is optimal}. This sounds reasonable but let's consider an example. Suppose there is a group of people and the task is to classify handwritten numbers. This is a difficult problem because the hand writing can be hard to read. For this reason one cannot expect that all people will achieve the optimal score but some people perform better than others. Hence, the behavior of every human is not optimal if compared to the maximal score or even the best performing human. Also, if we give the same group of people above a number of different tasks to solve then it is likely that not always the same person will perform best. Taken together, it doesn't make sense to make a computer behave like every human because most people do not perform optimal regardless what task we consider. So what it actually means is to make a computer perform like the best performing human. For one task this may actually mimic the behavior of one human, however, for several tasks this will correspond to a different human for every task. Hence, such a super human does not exist. That means if a machine can solve more than one task it doesn't make sense to compare it to one human because such a person does not exist, instead, it is compared to an ideal super human. For this reason, the answer to the above statement needs to be quanlified.

{\it When will the ultimate goals of AI be reached?} Over the years there have been a number predictions. For instance, Simon predicted in 1965 that "Machines will be capable, within twenty years, of doing any work a man can do" \cite{simon1965shape}, Minsky stated in 1967 that "Within a generation ... the problem of creating artificial intelligence will substantially be solved" \cite{minsky1967computation} and Kurzweil predicted in 2005 that strong AI, which he calls {\it singularity}, will be realized by 2045 \cite{kurzweil2005singularity}. Obviously, the former two predictions turned out to be wrong and the latter one is still in the future. However, predictions about undefined entities are anyway vague (see our discussion about intelligence above) and cannot be systematically evaluated.

From the above discussion one realizes that metaphors are frequently used in AI but they are not meant to be understood in a precise way but more as a motivation or stimulation. The origin of this might be related to the community behind AI which is considerably different from the more mathematics oriented communities in statistics or machine learning.

\section{Discussion}

\begin{figure}[t!]
\centering
\includegraphics[scale=0.75]{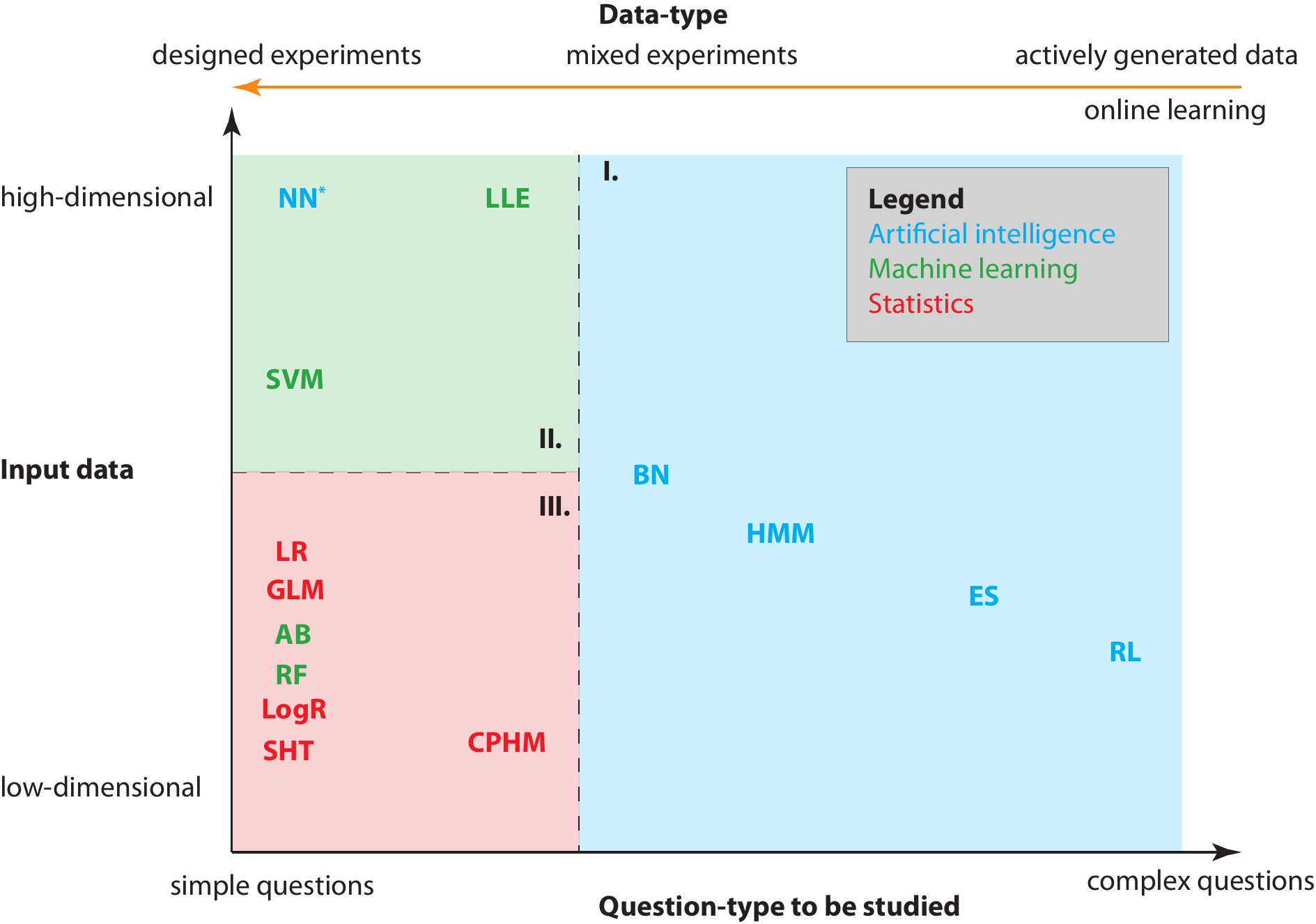}
\caption{A graphical overview of properties of core (and base) methods from artificial intelligence, machine learning and statistics. The x-axis indicates simple (left) and complex (right) questions a method can study whereas the y-axis indicates low- and high-dimensional methods. In addition, there is an orange axis (top) indicating different data-types. Overall, one can distinguish three regions where either methods form artificial intelligence (blue), machine learning (green) or statistics (red) dominate. 
}
\label{fig.ov}
\end{figure}

In order to obtain a general overview of the relations between methods in artificial intelligence, machine learning and statistics we shown a graphical overview of the properties of such methods in Fig. \ref{fig.ov}. The acronyms of the methods are given in Table \ref{tab.list}, listing core artificial intelligence, machine learning and statistics methods representing characteristic models. There are many properties of such methods, however, here we focus on two. Specifically, the x-axis indicates the question-type that can be addressed by a method from simple (left) to complex (right) questions, whereas the y-axis indicates the input dimensionality of the data from low- to high-dimensional. Overall, one can distinguish three regions where either methods from artificial intelligence (blue), machine learning (green) or statistics (red) dominate. Interestingly, before the introduction of deep learning neural networks region II. was entirely dominated by machine learning methods. For this reason we added a star to neural networks (NN) to indicate as a modern AI method. As one can see, methods from statistics are generally characterized by simple questions that can be studied in low-dimensional settings. Here by 'simple' we do not mean boring or uninteresting but rather 'specific' or 'well defined'. Hence, from Fig. \ref{fig.ov} one can conclude that AI tends to address complex questions that do not fit well into a conventional framework, e.g., as represented by statistics. The only exception are neural networks.

\begin{table}[ht]
  \small
  \begin{center}
    \caption{List of popular, core artificial intelligence, machine learning and statistics methods representing characteristic models of those fields.}
    \label{tab.list}
    \begin{tabular}{ l | c | c }
      \hline
      \textbf{Model} & \textbf{Application}  &  \textbf{References} \\
      \hline
      \hline
      \pbox{8cm}{Neural networks (NN)} & function approximation, classification  &  \pbox{7cm}{ \cite{rosenblatt1957perceptron,schmidhuber2015deep} } \\
      \hline
      \pbox{20cm}{Expert system (ES)}  & knowledge-based decisions &   \pbox{7cm}{ \cite{hayes1983building}} \\  \hline
      \pbox{8cm}{Hidden Markov models (HMM)} & sequential symbol processing  &  \pbox{7cm}{ \cite{rabiner1989tutorial} } \\
      \hline 
       \pbox{20cm}{Bayesian networks (BN)}  & uncertain reasoning &   \pbox{7cm}{\cite{pearl_1988,scutari_2010}} \\
       \hline
        \pbox{20cm}{Refinforcement learning (RL)}  & decision planing &   \pbox{7cm}{\cite{suttonbarto_1998,kaebeling_1996}} \\
      \hline \hline
 \pbox{20cm}{Support vector machines (SVM)}  & classification &   \pbox{7cm}{\cite{vapnik_1995,schoelkopf_smola_2002}} \\
      \hline   
 \pbox{20cm}{Adaptive Boosting (AB)}  & classification &   \pbox{7cm}{\cite{freund1997decision}} \\
      \hline        
 \pbox{20cm}{Locally Linear Embedding (LLE)}  & nonlinear dimensionality reduction &   \pbox{7cm}{\cite{roweis2000nonlinear}} \\
      \hline  
 \pbox{20cm}{Random forests (RF)}  & classification &   \pbox{7cm}{\cite{breiman_2001}} \\
      \hline  \hline      
 \pbox{20cm}{Linear regression (LR)}  & regression &   \pbox{7cm}{ \cite{weisberg2005applied,em_jmake1010021}} \\
      \hline  
       \pbox{20cm}{Logistic regression (LogR)}  & classification &   \pbox{7cm}{\cite{kleinbaum2002logistic}} \\
      \hline 
 \pbox{20cm}{Generalized linear models (GLM)}  & regression &   \pbox{7cm}{\cite{nelder1972generalized,dunn2018generalized}} \\
      \hline  
 \pbox{20cm}{Statistical hypothesis testing (SHT)}  & hypothesis testing &   \pbox{7cm}{\cite{sheskin_2004,em_jmake1030054}} \\
      \hline 
 \pbox{20cm}{Cox proportional hazard model (CPHM)}  & survival analysis &   \pbox{7cm}{\cite{cox1972regression,kleinbaum_2005}} \\
      \hline \hline       
    \end{tabular}
  \end{center}
\end{table}

For most of the methods shown in Table \ref{tab.list} there exist extensions to the 'base' method. For instance, a classical statistical hypothesis testing is conducted just once. However, modern problems in genomics or the social sciences require the testing of thousands or more hypotheses. For this reason multiple testing corrections have been introduced \cite{farcomeni_2008,em_jmake1020039}. Similar extends can be found for regression. However, if only the original methods are considered one obtains a simplified categorization for the domains of AI, ML and statistics. 
\begin{itemize}
\item Traditional domain of artificial intelligence $\Rightarrow$ Complex questions
\item Traditional domain of machine learning $\Rightarrow$ High-dimensional data

\item Traditional domain of statistics $\Rightarrow$ Simple questions
\end{itemize}

In Fig. \ref{fig.ov}, we added one additional axis (orange) on top of the figure indicating different data-types. In contrast to the axes for the question-type and the input data, the scale of this axis is discrete that means there is no smooth transition between the corresponding categories. Using this as an additional perspective one can see that machine learning as well as statistics methods require data from designed experiments. This form of experiment corresponds to the conventional types of experiments in physics or biology where the whole measurement follows a predefined plan (experimental design). In contrast, AI methods use frequently actively generated data (also known as online learning). An example for this data type is the data a robot generates exploring its environment or data corresponding to moves in a games.

We think it is important to emphasize that (neither) methods from AI (nor from machine learning or statistics) cannot be mathematically derived from a common, underlying methodological framework but they have been introduced separately and independently. In contrast, physical theories, e.g., about statistical mechanics or quantum mechanics, can be derived from a Hamiltonian formalism or alternatively from Fisher Information \cite{goldstein2013classical,frieden1998physics}.

Maybe the most interesting insight from Fig. \ref{fig.ov} is that the currently most successful AI methods, namely neural networks, do not address complex questions but simple ones (e.g., classification or regression) for high-dimensional data. This is notable because it goes counter the tradition of AI taking on novel and complex problems. Also considering the current interest in futuristic problems, e.g., self-driving cars, automatic trading or health diagnostics this seems even more curious because it means such complex questions are addressed reductionistically  dissecting the original problem into smaller subproblems rather than addressing them as a whole. Metaphorically, this may be considered as maturing process of AI settling after a rebellious adolescence against the limitations of existing fields like control theory, signal processing or statistics \cite{russell2016artificial}. If it stays in this way remains to be seen in the future.

Finally, if one considers novel extensions for all base methods from AI, ML and statistics one can summarize the current state of these fields as follows.
\begin{itemize}
\item Current domain of artificial intelligence, machine learning and statistics $\Rightarrow$ Simple questions for high-dimensional data
\end{itemize}

\section{Conclusions}

In this paper, we discussed the desired and current state of AI and clarified its goals. Furthermore, we put AI into perspective to machine learning and statistics and identified similarities and differences. The most important results can be summarized as follows:
\begin{enumerate}
\item Currently, no generally accepted definition of 'intelligence' is available. $\Rightarrow$ AI if mathematically undefined, almost 65 years after its formal inception.
\item The aspirations of AI are very high focusing on ambitious goals. $\Rightarrow$ AI is not explicitly data focused - in contrast to data science.
\item General AI methods do not provide neurobiological models of brain functions. $\Rightarrow$ AI methods are merely means to analyze data - similar to methods from machine learning and statistics.
\item Supplement: Also deep neural networks do not provide neurobiological models of brain functions. $\Rightarrow$ They are merely means to analyze data.
\item The currently most successful AI methods, i.e., deep neural networks, focus on simple questions (classification, regression) and high-dimensional data. $\Rightarrow$ This goes counter traditional AI but is similar to contemporary machine learning and statistics.
\item AI methods are not derived form a common mathematical formalism but have been introduced separately and independently. $\Rightarrow$ There is no common conceptual framework that would unite the ideas behind the different AI methods. 
\end{enumerate}

The closeness to applications of AI is certainly good for making it practically relevant and achieving an impact in the real world. Interestingly, this is not unalike to a commercial product. A downside is that AI comes also with slogans and straplines used for marketing reasons just as for regular commercial products. We hope our article can help locking behind the marketing curtain of AI to see what the field is actually about from a scientific perspective.

\section*{Conflict of Interest Statement}

The authors declare that the research was conducted in the absence of any commercial or financial relationships that could be construed as a potential conflict of interest.

\section*{Author Contributions}

All authors contributed to all aspects of the preparation and the writing of the manuscript.



\bibliographystyle{unsrt}

\bibliography{/Users/fes/fes/Artikel/Bib_ref/ref_list,/Users/fes/Local/Mails/CV/Bibref/bib_ref_journals,/Users/fes/Local/Mails/CV/Bibref/bib_ref_books_editor,/Users/fes/Local/Mails/CV/Bibref/bib_ref_book_chapters,/Users/fes/Local/Mails/CV/Bibref/bib_ref_thesis}

\end{document}